\documentclass[letterpaper, 10 pt, conference]{ieeeconf} 
\IEEEoverridecommandlockouts                   
\usepackage{cite}
\usepackage[colorlinks=true,citecolor=green]{hyperref}
\usepackage{amsmath,amssymb,amsfonts}
\usepackage{algorithmic}
\usepackage{graphicx}
\usepackage{booktabs}
\usepackage{float}

\title{Robo-GS: A Physics Consistent Spatial-Temporal Model for Robotic Arm with Hybrid Representation\\
}

\makeatother
\author{Haozhe Lou$^{1*}$, Yurong Liu$^{2*}$, Yike Pan$^{3*}$, Yiran Geng$^{4*}$, Jianteng Chen$^{5*}$, Wenlong Ma$^{6}$,\\Chenglong Li$^{6}$, Lin Wang$^{6}$, Hengzhen Feng$^{6}$, Lu Shi$^{9}$, Liyi Luo$^{8}$, Yongliang Shi$^{\dag 7}$
\thanks{$^{*}$Equal contribution, $^\dag$Corresponding author.}
\thanks{$^{1}$University of Southern California, $^{2}$National University of Singapore, $^{3}$University of Michigan, Ann Arber,$^{4}$ Peking University, $^{5}$Hong Kong University of Science and Technology, $^{6}$Beijing Institute of Technology, $^{7}$Tsinghua University, $^{8}$Xiaomi Robotics Lab, $^{9}$AIR, Tsinghua University.}}%
\begin{document}

\maketitle
\thispagestyle{empty}
\pagestyle{empty}


\begin{abstract}
The Real2Sim2Real (R2S2R) paradigm is critical for advancing robotic learning. Existing methods lack a comprehensive solution to accurately reconstruct real-world objects with both spatial representations and their associated physics attributes in the Real2Sim stage.

We propose a Real2Sim pipeline to generate digital assets enabling high-fidelity simulation. We design a hybrid representation model that integrates mesh geometry, 3D Gaussian kernels, and physics attributes to enhance the representation of robotic arms in digital assets.
This hybrid representation is implemented through a Gaussian-Mesh-Pixel binding technique, which establishes an isomorphic mapping between mesh vertices and the Gaussian model. This enables a fully differentiable rendering pipeline that can be optimized through numerical solvers, achieves high-fidelity rendering via Gaussian Splatting, and facilitates physically plausible simulation of the robotic arm's interaction with its environment through mesh geometry.
With the digital assets, we propose a fully manipulable Real2Sim pipeline that standardizes coordinate systems and scales, ensuring the seamless integration of multiple components. To demonstrate its effectiveness, we include datasets covering various robotic manipulation tasks with their mesh reconstructions. Our model achieves state-of-the-art results in realistic rendering and mesh reconstruction quality for robotic applications. Our code and datasets will be made publicly available at \href{https://robostudioapp.com/}{robostudioapp.com}.
\end{abstract}


\section{Introduction}

Real2Sim2Real plays a critical role in robotic arm control and reinforcement learning but remains a challenging task due to the complex physical properties of robots and the objects they manipulate. It tackles well-known problems of existing learning frameworks in robot autonomy such as model generalization ability and label availability~\cite{c24}. Still, it has yet to be fully realized due to significant challenges in spatial and color representation as well as rendering quality within current Real2Sim approaches~\cite{c2}. These challenges hinder transferring learned policies from simulation to real-world applications and compromise the reliability and performance of robotic systems trained in simulated environments. 

\begin{figure}[!t]
\vspace{-6mm}
\centering
\includegraphics[width=0.46\textwidth]{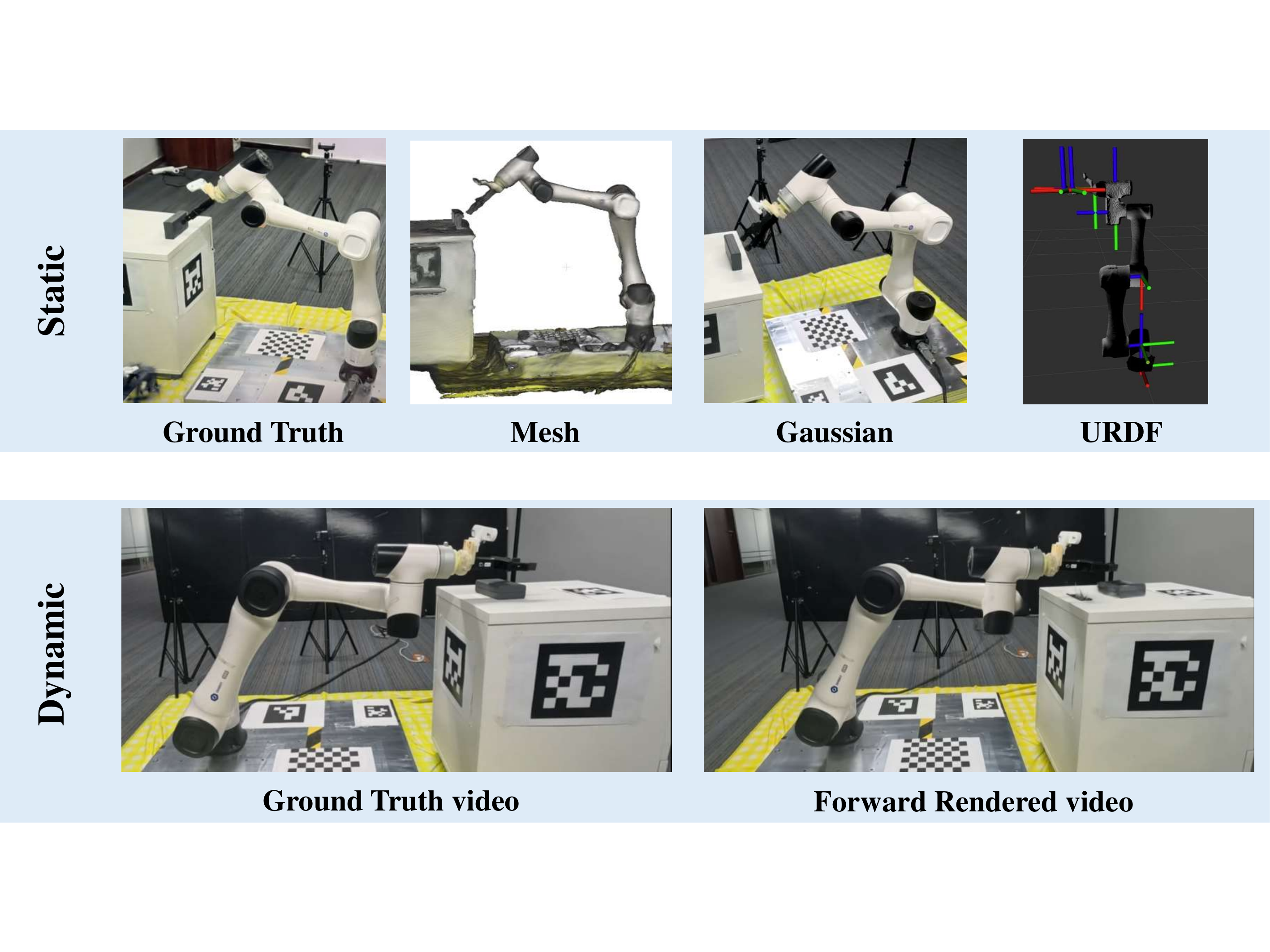}%
\vspace{-9mm}
\caption{The reconstructed digital assets include the extracted mesh, Gaussian Models~\cite{c1}, a dynamically consistent kinematic model.}
\label{assetproduction}
\vspace{-5mm}
\end{figure}

In this paper, we target the Real2Sim and demonstrate a \textbf{holistic reconstruction} of robotic arm operation scenes, which requires a (i) manipulable robot model, (ii) reconstructed background and objects, (iii) physical parameters such as mass and friction. Our approach employs Unified Robot Description Format(URDF, which is a standard XML format used to describe the physical configuration, joints, and kinematic structure of robots)~\cite{c3,c4}, as a spatial representation and integrates governing equations with physics parameters as the forward deformation mechanism. This combination enables accurate collision detection and consistent rendering across both simulation and Gaussian Splatting (GS) environments~\cite{c5,c6}. 


The core to our method is a \textbf{Gaussian-Mesh-Pixel binding}, which establishes an isomorphic relationship between mesh vertices, Gaussian kernels~\cite{c1}, and image pixels. Each Gaussian is assigned a semantic label and a corresponding ID, enabling the precise application of transformation matrices governed by the URDF. This ensures a seamless transfer of trajectories between real-world video, simulation results, and rendered images. The advantages of this binding include (i) End-to-end differentiable gradient passing between each representation, allowing backward optimization to be performed across different backends; (ii) Superior collision detection through our state-of-the-art mesh reconstruction; and (iii) High rendering quality. 


Our system ensures a faithful Real2Sim process, allowing learned policies to be effectively deployed in real-world scenarios (Please see the supplementary video). Additionally, it supports editing within the robotic simulator Isaac Sim (Gym)~\cite{c6,c7} backend, enabling novel-pose and novel-policy adjustments. Optimized for robotic arms in CR3, CR5~\cite{c8} and UR5~\cite{c9} product sequences, our method is versatile enough to generalize to other robotic arm models. Our method achieves state-of-the-art performance in mesh reconstruction and dynamic rendering compared to current methods~\cite{c10}.

Furthermore, we propose a new format for digital assets, represented by a combination of mesh, Gaussian Splatting, and real-world motion~\cite{c11}, as shown in Fig.\ref{assetproduction}. This approach surpasses traditional textured mesh and material attributes by integrating critical physics parameters, such as mass and friction, extracted from real-world motion videos~\cite{c12}.

\begin{itemize}
    \item We introduce a holistic framework for robotic arm control and simulation, combining mesh geometry, 3D Gaussian kernels, and physics attributes to achieve high-fidelity scene reconstruction and physically accurate simulation of robotic manipulation tasks. Our method, based on a Gaussian-Mesh-Pixel binding technique, ensures seamless integration between simulation, rendering, and real-world data, allowing for precise policy training and control optimization in complex environments.

    \item We address the challenges of accurately reconstructing robotic arm movements, including occlusions and inconsistencies between simulated and real-world data.

    \item Experiments and ablations demonstrate the effectiveness of our method. It achieves state-of-the-art performance in mesh reconstruction and high-quality rendering, significantly outperforming previous approaches in both static and dynamic scenarios.
\end{itemize}

\begin{figure}[t]
\centering
\includegraphics[width=0.48\textwidth]{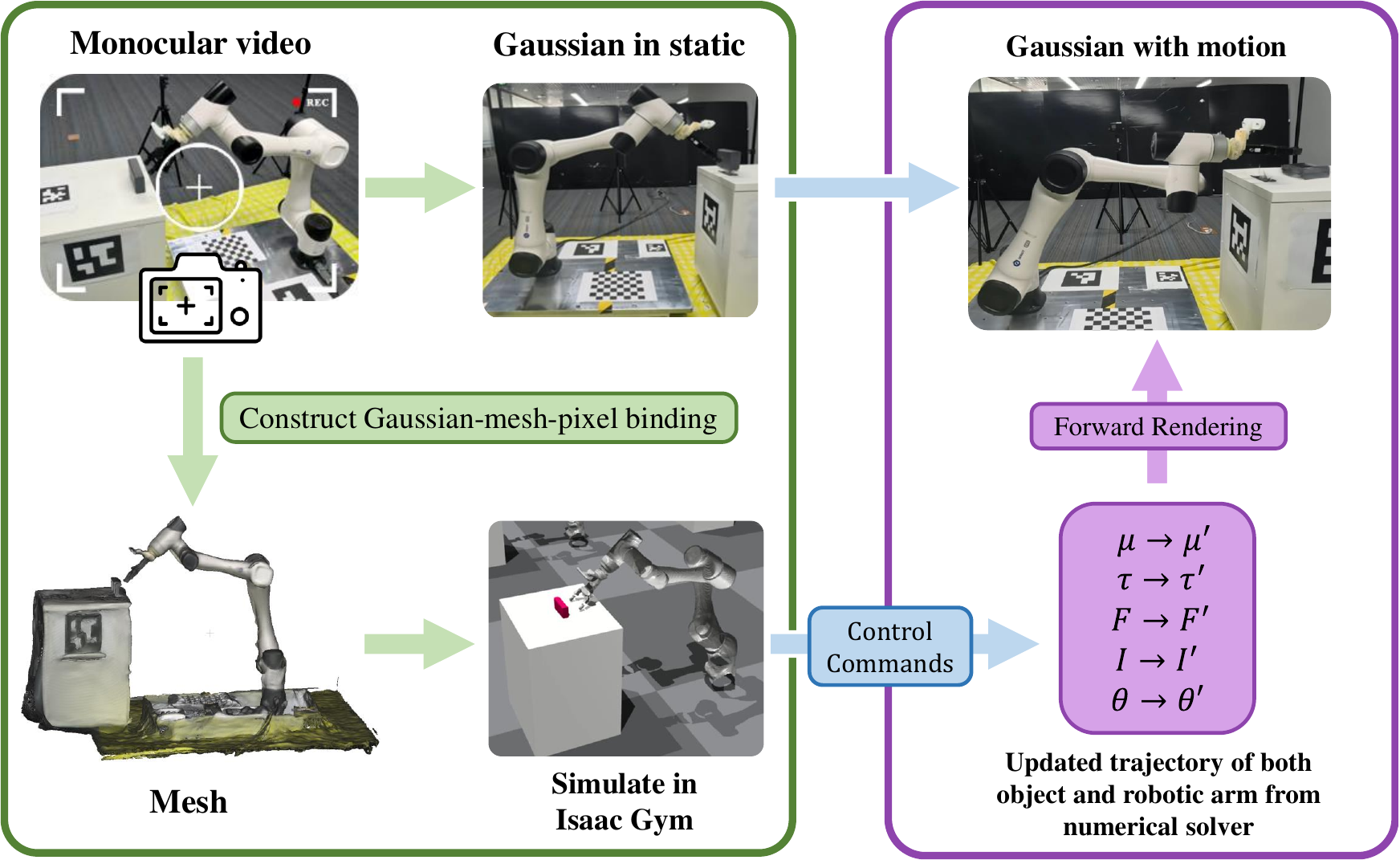}%
\vspace{-6mm}
\caption{(Left) Our method converts monocular video to 3D meshes and Gaussians, URDF models, and generates trajectories through Gaussian-Mesh-Pixel binding, and (Right) renders dynamic interactions using forward deformation based on updated object and robotic arm trajectories.}
\label{system}
\vspace{-2mm}
\end{figure}

\section{related work}

\subsubsection*{Rendering}
The Real2Sim2Real paradigm, crucial for bridging physical and simulated environments, relies on advanced rendering techniques. Recent innovations in 3D reconstruction, such as Gaussian Splatting and Neural Radiance Fields (NeRF)~\cite{c1,c13,c14,c15,c16}, have significantly enhanced the creation of high-fidelity digital twins, essential for accurate simulations.

Robotic simulation has progressed with the use of transformation matrices and Linear Blend Skinning (LBS)~\cite{c17} control modes. However, these methods often struggle with maintaining geometric consistency in robotic arms due to rigid mesh motions. SC-GS~\cite{c18} addresses this by employing sparse control points, though it falls short in accurately simulating robotic arm movements.

Further advancements include the use of NeRFs and Gaussian Splatting for reconstructing robotic operation scenes~\cite{c19,c10,c20}. MD-Splatting~\cite{c21} explores Gaussian-based reconstruction of highly deformable objects, utilizing NeRF and particle-based representations. However, these methods often lack high-fidelity rendering capabilities~\cite{c22}. Our approach enhances 4D Gaussian Splatting quality by replacing traditional MLP-based deformation fields~\cite{c18,c23,c72} with numerical ODE solvers~\cite{c24,c25}.

\subsubsection*{Asset Extraction}
Significant progress in mesh reconstruction and simulation integration has been made with NeRFmeshing~\cite{c26} and SUGAR~\cite{c27,c28}. NeRFmeshing utilizes NeRFs for mesh reconstruction, while SUGAR employs Gaussian Splatting for mesh extraction. However, the complex and sparse nature of robotic arm structures often leads to non-smooth reconstructions due to challenges in normal estimation.
The 2DGS method~\cite{c30} effectively addresses non-smooth surface issues by introducing a planar-like 2DGS representation. After applying cleaning and matching techniques, the quality of meshes generated by 2DGS proves sufficient for robotic simulation purposes.

\subsubsection*{Control}
In the Real2Sim2Real paradigm, robotic arm control typically employs pose-based methods, transferring end-effector pose trajectories~\cite{c31} between real-world and simulation policies. This approach uses inverse kinematics to derive pre-joint control information~\cite{c32}. Current control policies include diffusion-based models~\cite{c33} and trajectory-to-video generation methods like IRASim~\cite{c34}, which utilize generative methods driven by trajectory for simulation~\cite{c36} and representation. However, these methods are limited in fully representing 3D and 4D real-world scenarios~\cite{c35, c39}. Traditional robotic arm control connects linkages with predefined joints, using angles for path control~\cite{c41,c42,c43}. In a Gaussian setting, where each Gaussian is treated as an independent element~\cite{c37}, this approach can lead to motion inconsistencies. Our method addresses this issue through isomorphism Mesh-Gaussian binding.

Recent innovations include a signed distance field-based approach for robotic arm morphology~\cite{c44}, using joint angles as input with a neural decoder guiding the deformation field. However, this method lacks explicit compatibility with the MOVEIT path-based control pipeline~\cite{c45} and current reinforcement training toolkits~\cite{c41,c46} UMI on Legs demonstrates the potential of simulation-based manipulation tasks, though scene and object reconstruction remains a challenge~\cite{c47}.

Physically embodied Gaussian Splatting~\cite{c48} employs Gaussian distributions and particles derived from pre-built meshes to generate movement~\cite{c38}, tested on simulated data. Our approach advances this concept by utilizing meshes extracted from video data for more accurate representation, coupled with real-world scenario data collection.

\section{method}



\subsection{Representation}

Our digital asset is represented by a mesh, a set of Gaussian primitives, and the spatial-temporal trajectory of the assets. Traditionally, digital asset production focuses on the textured mesh and material attributes~\cite{c63}. 
However, physical parameters such as mass and friction coefficients are necessary to enable Sim2Real policy training for downstream robotic manipulation tasks. Thus we incorporate these physical parameters in our representation.

A polygon mesh $M = (V, F)$ is defined by a set of vertices $V = \{\nu_1, \nu_2, \dots, \nu_n\}, \nu_i\in\mathbb{R}^3$ and faces $F = \{f_1, f_2, \dots, f_k\}$.
Similar to 3DGS~\cite{c1}, we use a set of 3D Gaussian primitives $\{G\}_{p \in P}$, where each Gaussian primitive $G$ includes a center $\textbf{x}_p$, an opacity $\sigma_p$, a covariance matrix $C_p$, a set of spherical harmonic (SH) coefficients $SH_p$, and a semantic category of the Gaussians $S_p$, respectively~\cite{c49}.

We here describe the proposed Gaussian-Mesh-Pixel binding (Fig.~\ref{GMP}): 
We first define a Gaussian-Mesh mapping $\mu$ that associates each Gaussian primitive with a set of mesh vertices (and faces). Second, we project the Gaussian primitives into 2D via projection mapping, thus establishing an isomorphism between pixel locations $P$ and Gaussian kernels $G$. By compositing the above mappings, we define an isomorphic binding between Gaussians, mesh, and pixels. Here we define each mapping concretely:

\begin{figure}[!t]
\centering
\includegraphics[width=0.45\textwidth]{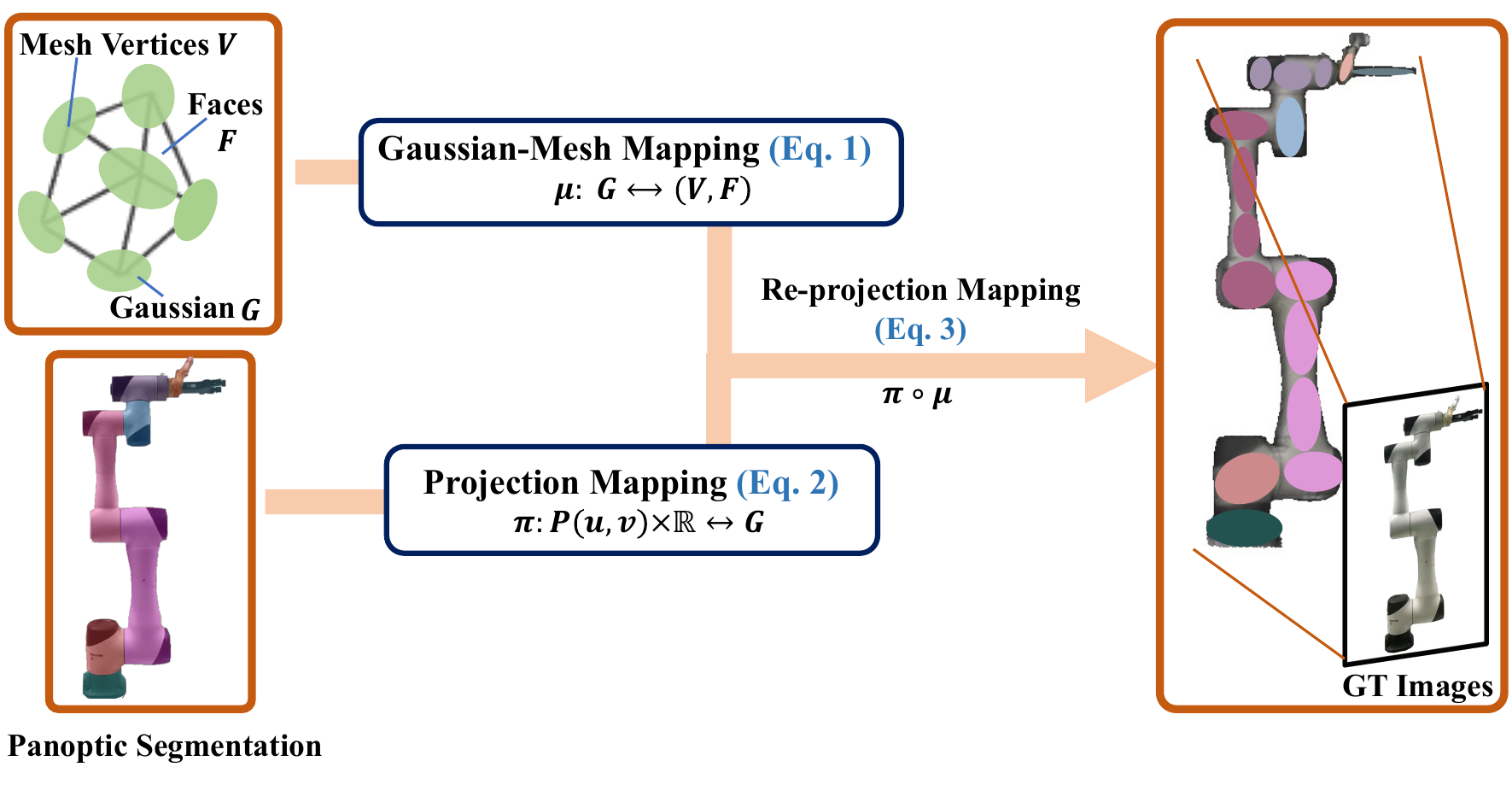}%
\caption{Gaussian-Mesh-Pixel binding: This binding preserves the structural properties between Gaussian $G$, mesh $(V, F)$, and pixel location $P$ under affine transformation, which is a composition (right) a Gaussian-Mesh mapping (top-left), and a projective Pixel-Gaussian binding (top-right).}
\label{GMP}
\vspace{-5mm}
\end{figure}

\begin{enumerate}
  \item \textbf{Gaussian-Mesh Mapping:} An order-preserving mapping \( \mu: \mathbb{R}^3 \to V \) associates \( A \) with a vertex in \(V\):
  \begin{equation}
  \mu(G(x, y, z)) = \{ \nu_{n}, f_{k} \vert \nu_{n} \in V, f_{k} \in F \}.
  \end{equation}
  
  \item \textbf{Projection Mapping:} Given a known camera pose represented in $\mathbb{SE}(3)$ and camera intrinsic, any known 3D point location can be re-projected to the 2D image plane using the perspective projection model: 
  \begin{equation}
  (u, v) = \pi(G(x, y, z)).
  \end{equation}

  \item \textbf{Re-projection Mapping:} Through the composition of the above mappings, we define an isomorphic relationship \( \phi \) that associates Gaussian \( G \) with both pixel location \( P \) and vertices \( V \)\cite{c5}:
  \begin{equation}
  \phi(G(x, y, z)) = (\pi \circ \mu)(G(x, y, z)).
  \end{equation}
\end{enumerate}

This mapping connects the trajectory between real-world scenes captured on the image plane, simulation results from the mesh-based engine, and rendered scenes~\cite{c50,c51,c52}. It unifies the representations of Gaussian Splatting, mesh, and real-world motion, ensuring spatio-temporal consistency and allowing gradients to flow from real-world video to Gaussian to mesh (backward optimization) and from mesh to Gaussian to rendered video (forward rendering).

\subsection{Mesh extraction}
To obtain necessary geometric cues from the video sequences, we first collect a set of static images of the scene and utilize Gaussian Splatting techniques to prepare the meshes.
We extract linkage, object, and background mesh through our mesh extraction and cleaning technique. Specifically, we adapt surfel-based Gaussian Splatting~\cite{c30}, along with vertex cleaning and our re-orientation strategy.

We follow 2DGS~\cite{c30} and represent a Gaussian with a ``2D'' surfel rather than a full 3D Gaussian kernel, this is characterized by a center vector $\textbf{x}_p$, a tangent vector $\textbf{t}_p = (\textbf{t}_u, \textbf{t}_v)$, and scaling $\textbf{s}_p = (s_u, s_v)$. The normal vector of the Gaussian primitive is thereby defined as $\mathbf{N}_w = \mathbf{t}_u \times \mathbf{t}_v$. A Gaussian surfel is then defined in a local tangent plane as:
\begin{equation}
\mathbf{x}_p(u, v) = \mathbf{x}_p + s_u \mathbf{t}_u u + s_v \mathbf{t}_v v = \mathbf{H}(u, v, 1)^\mathrm{T}
\end{equation}
\begin{equation}
\mathbf{H} =
\begin{bmatrix}
s_u \mathbf{t}_u & s_v \mathbf{t}_v & \mathbf{0} & \mathbf{p}_k \\
0 & 0 & 0 & 1
\end{bmatrix}
=
\begin{bmatrix}
\mathbf{RS} & \mathbf{p}_k \\
0 & 1
\end{bmatrix}
\end{equation}
where $\mathbf{H} \in \mathbb{R}^{4 \times 4}$ is a homogeneous transformation matrix representing the geometry of the Gaussian surfel. For the point $\mathbf{u} = (u, v)$ in $uv$ space, its probability density function can then be expressed as: $\mathcal{G}(\mathbf{u}) = \exp \left( -\frac{u^2 + v^2}{2} \right)$.

After the reconstruction process, we can extract a mesh based on the Gaussian kernels using TSDF fusion~\cite{c54}. However, there are discrepancies between the extracted meshes and the requirements for URDF production.

We identify four major issues and propose solutions for each to achieve alignment between the real-world robotic arm and Gaussian Splatting.

\begin{itemize}
\item \textbf{Define a Unified Coordinate System and Origin:} A unified coordinate system integrates real-world and simulation scenes using the OpenGL y-up axis, which is consistent in both Gaussian Splatting and the simulation engine. A reference point, such as the base of the robotic arm, must be set as the origin in Gaussian Splatting to ensure a consistent coordinate definition.
\item \textbf{Scale Adjustment:} To compensate for discrepancies between the reconstructed scene and the real world—inevitable when reconstructing from monocular videos but essential to address for accurate robotic control and simulation—we manually correct them by establishing a standard based on real-world measurements using MeshLab.
\item \textbf{Orientation Alignment:} The reconstructed mesh from Gaussian Splatting may not be axis-aligned due to the settings of Structure-from-Motion (SfM)~\cite{c55}. To correct this, we use a scene re-orientation technique to align the reconstructed scenes with the simulation engine's coordinate system~\cite{c16}. This alignment is crucial for accurately reproducing simulated movements and interactions in the real world.
\item \textbf{Generate Physics Parameters:} The URDF requires mass, friction, and damping factors for physics simulation. We embed the LLM-inferred physics parameters and governing equation categories into the URDF asset, based on Panoptic information~\cite{c29,c58}.

\end{itemize}

\subsection{Kinematic Governing Equation}



\begin{figure}[!t]
\centering
\includegraphics[width=0.48\textwidth]{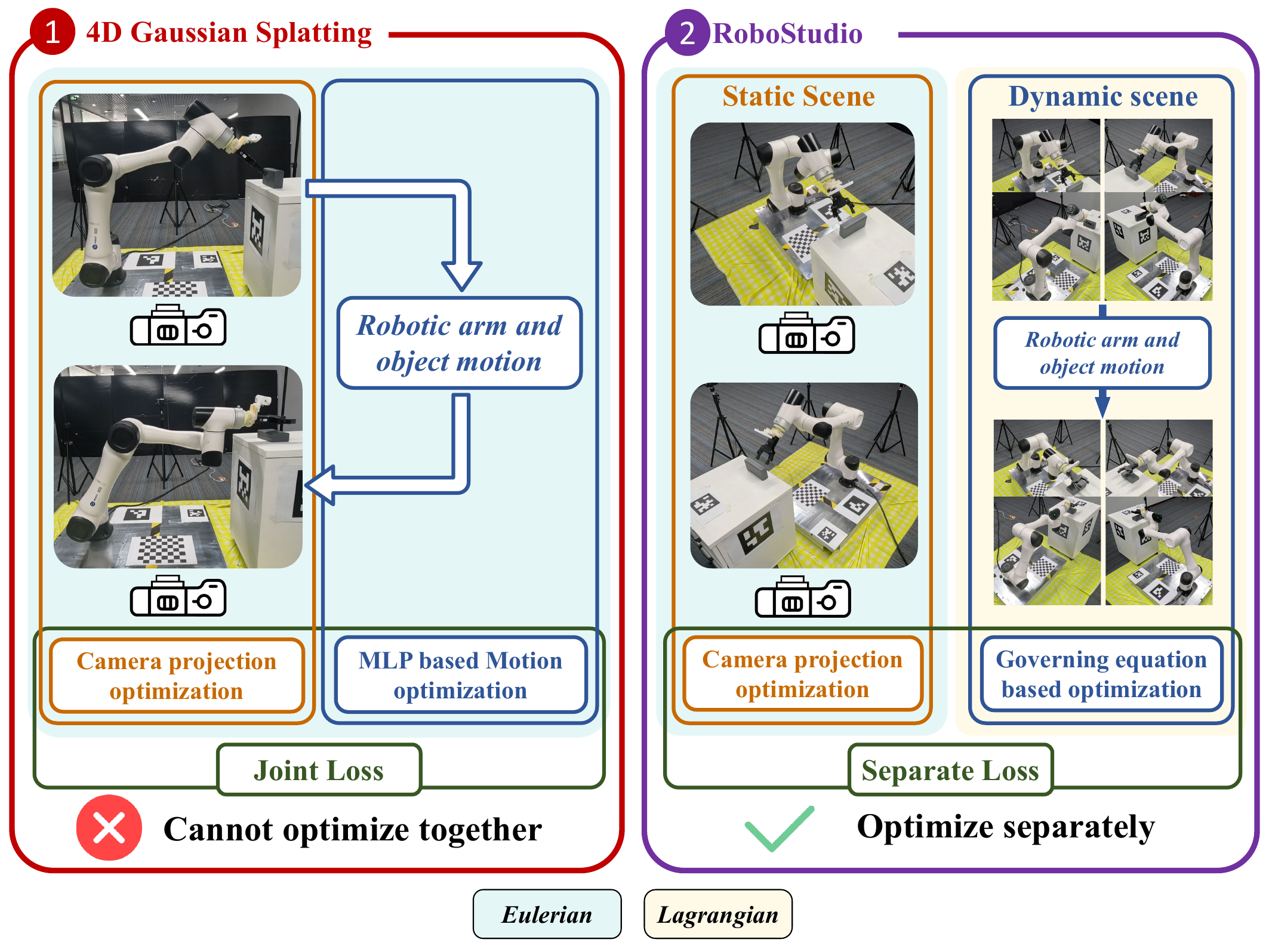}%
\vspace{-5mm}
\caption{Comparison between deformable Gaussian Splatting (left) and our solution. (Left) In general deformation-based Gaussian Splatting~\cite{c18,c10}, the dynamics of the scene are embedded in each Gaussian primitive, resulting in a high-dimensional degree of freedom (DoF). This causes failure during the training process in complex dynamic scenes, as observed in our setting. (Right) Our approach decomposes the motion into a small number of linkages and objects driven by governing equations, significantly reducing the degrees of freedom. This allows numerical solvers to optimize more efficiently.}
\label{optimization_result}
\vspace{-2mm}
\end{figure}

Optimizing 3DGS involves minimizing the re-projection error between the Gaussian representation and pixel data. In 4D Gaussian Splatting, input images include both pixel data and timestamps, aiming to optimize the 4D reconstruction under the XYZT representation. This non-convex problem is challenging due to the lack of multi-view consistency in this setting, as shown in Fig. \ref{optimization_result}.

We find that the optimal solution to this ill-posed problem in robotic scenes is to decompose the original 4D reconstruction problem into a two-stage process: static and dynamic stages. We treat static scenes using the Eulerian representation and dynamic scenes using the Lagrangian representation~\cite{c51}. The static stage aims to optimize the camera reprojection loss, while the dynamic stage focuses on simulating the motion of objects and the robotic arm based on their corresponding governing equations. Through our Mesh-Pixel-Gaussian binding approach, we can pass gradients and accurately simulate the results.

In our system, control information is aligned with a predefined governing equation for physics simulation and rendering~\cite{c40}. However, when simulating these scenes in a Gaussian Splatting framework, the joint angle based trajectory representation are incompatible with the discrete Gaussian setting because the spatial-temporal correspondence between joints and Gaussians cannot be accurately determined~\cite{c44}. To overcome this, we define a kinematic and dynamic system tailored for Gaussian-based robotic arm and gripper pose control, focusing on controlling the end-effector's pose and using inverse kinematics to generate control signals for each joint in the real world.

The kinematics of the robotic arm are adapted to Gaussian Splatting using Modified Denavit-Hartenberg (MDH) parameters and coordinate transformations across Gaussian Splatting, simulation, and real-world settings. A point on link \(i\) in Cartesian coordinates can be described using the MDH parameters~\cite{c56}, which include link length ($a_i$), link twist ($\alpha_i$), link offset ($d_i$), and joint angle ($\theta_i$). These parameters define the forward kinematic transformation matrix $T_i$ between consecutive coordinate frames, allowing the transformation from link $i-1$ to link $i$.




\begin{equation}
T_{i-1}^i = R_x(\theta_{i-1}) T_x(a_{i-1}) R_z(\alpha_i) T_z(d_{i})
\end{equation}

\begin{equation}
T_i =
\begin{bmatrix}
\cos \theta_i & -\sin \theta_i & 0 & a_i \\
\sin \theta_i \cos \alpha_i & \cos \theta_i \cos \alpha_i & -\sin \alpha_i & -\sin \alpha_i d_i \\
\sin \theta_i \sin \alpha_i & \cos \theta_i \sin \alpha_i & \cos \alpha_i & \cos \alpha_i d_i \\
0 & 0 & 0 & 1
\end{bmatrix}
\end{equation}
The value of MDH parameter can be supplied from either CAD designing, real-world measuring, or manual values.

In this approach, the mesh serves as an intermediary for mapping movements. The motion of each mesh is transferred to the Gaussian bound to it, guiding the movement of the Gaussian~\cite{c53,c57}. Unlike traditional inverse kinematic control method~\cite{c74}, which represents trajectories under continuous space, we model the system as a discrete state space. In this model, each timestamp is independent but shares a common coordinate system (Fig. \ref{optimization_result}). To account for deformation between timestamp \(0\) and timestamp \(t\), we first map the point \(P_0 = (x_0, y_0, z_0)\) to its base coordinate \({P}_{\text{base}}\):


\begin{equation}
\mathbf{P}_{\text{base}} = (\mathbf{T}_0^i)^{-1} \mathbf{P_0}
\end{equation}
Then, map the point forward to the new frame at time \(t\), \(P_t = (x_t, y_t, z_t)\):

\begin{equation}
\mathbf{P}_{t} = \mathbf{T}^i_{0_{t}} \mathbf{P}_{\text{base}}
\end{equation}
The overall transformation from the initial frame to the new frame at time \(t_1\) is:

\begin{equation}
\mathbf{P}_{t1} = \mathbf{T}^i_{0_{t}} (\mathbf{T}_0^i)^{-1} \mathbf{P_0}
\end{equation}

Since the initial position of the robotic arm is the default position, we can simply load the trajectory from the policy and transform it into the mesh and Gaussian kernels form.






\subsection{Dynamic Governing Equation}\label{sec:method_dynamic}
The dynamics of the rigid asset are governed by the Newton-Euler equations. The governing equation transforms the optimization problem from 6n degrees of freedom (DOF) to 6A DOF, where A is the number of movable parts defined by instance segmentation, and n is the number of Gaussian~\cite{c58,c59}. This can simplify the representation and optimization of the numerical solver.

Following~\cite{c50,c52}, we define the Newton-Euler equations as

\begin{equation}
\begin{bmatrix}
m\mathbf{I}_{3\times3} & \mathbf{0} \\
\mathbf{0} & \mathbf{I}
\end{bmatrix}
\begin{bmatrix}
\dot{\mathbf{v}} \\
\dot{\boldsymbol{\omega}}
\end{bmatrix}
=
\begin{bmatrix}
\mathbf{f} \\
\boldsymbol{\tau}
\end{bmatrix}
-
\begin{bmatrix}
\mathbf{0} \\
\boldsymbol{\omega} \times \mathbf{I}\boldsymbol{\omega}
\end{bmatrix}
\end{equation}
The simulation of forces, inertial, and torque are performed in a physics simulator based on our generated mesh and inferences parameter. We simply perform the semi-implicit Euler integration from Gradsim~\cite{c50} for numerical solving. Dynamic governing equation generates the transformation matrix of force based control of robotic arm and the motion of rigid object. The simulation of soft body can follow PAC-nerf, PhysGaussian and Simplicits~\cite{c51,c57,c60}. 

We decompose the Gaussian movement into a global transformation matrix under the supervision of kinematics and dynamics. Unlike the deformation-based Gaussian Splatting~\cite{c18,c62}, which treats movement as an MLP deformation field, our approach explicitly updates the position of Gaussians in each timestamp and is particularly effective for objects with rigid and linked motion.

To render a view, Gaussian Splatting projects these 3D Gaussians onto the image plane. The final color of each pixel is computed as~\cite{c51,c57}
\begin{equation}
C = \sum_{k \in P} \alpha_k SH(d_k; C_k) \prod_{j=1}^{k-1} (1 - \alpha_j). \label{eq:final_color}
\end{equation}
Here, $\alpha_k$ represents the z-depth ordered effective opacity, i.e., products of the Gaussian weights and their overall opacity $\sigma_k$; $d_k$ stands for the view direction from the camera to $x_k$. We only compute transformation for the $k_{th}$ Gaussians that shares same panoptic mask with this pixel.

For the effect of global transformation matrix \(T^{sam}_p\) apply on static scenes, we decompose it into the effect of Rotation and translation:

\begin{equation}
f_t(d) = f_0(R^T d).
 \end{equation}
The equation \(f_t(d)\) represents the effect of rotation on the view direction of Gaussian~\cite{c57}. We define \(\tau\) to represent the position update of each Gaussian center at a new time $t$, according to a transformation conditioned on the kinematics and dynamics of the moving part,
\begin{equation}
  \tilde{\tau}_p(X, t, S) = R^{S}_{t}\cdot{x^{S}_p} +T^{S}_{t}.
\end{equation}
where S indicates the instance ID of the moving part.

\section{Experiment}

\subsection{Datasets:}
Current datasets~\cite{c64,c65,c66,c67} contain only videos, language commands, and robot trajectories. Building digital assets based on these datasets would result in a significant Real2Sim gap~\cite{c64}. To address this gap, we propose that a comprehensive representation for a Real2Sim dataset should include monocular video with panoptic labeling, bounding box annotations, physics parameters, mesh geometry, policy, and trajectory information~\cite{c68,c69}, and real-world motion data of both the object and the robotic arm during policy implementation, along with URDF files for CR and UR robotic arm sequences. We find that this dataset format provides all the necessary information for policy training while ensuring millimeter and millisecond-level precision.

\subsection{Rendering Results:}
\begin{figure*}[ht]
\centering
\includegraphics[width=0.96\textwidth, height=0.19\textheight]{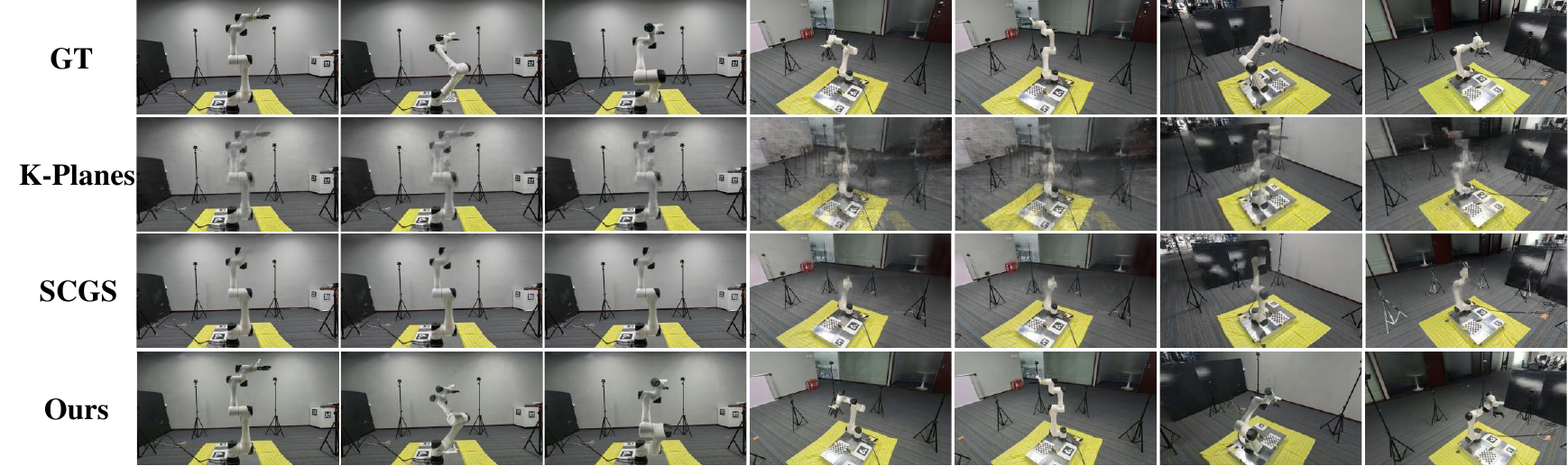}%
\caption{Qualitative result for Novel-pose}
\label{novelpose}
\end{figure*}

We compare the current state-of-the-art in 4D Gaussian Splatting and 4D-NeRF, including SC-GS~\cite{c18} and K-Planes~\cite{c70}, with our method. Fig. \ref{novelpose} illustrates the reconstruction of a robotic arm performing a series of trajectories. We observe that both K-Planes and SC-GS struggle to accurately optimize the transformation of the robotic arm and object motion, resulting in poor rendering outcomes. As discussed in section \ref{sec:method_dynamic}, unlike SC-GS and K-Planes, which optimize the representation across the entire space, our method simplifies the dynamic Gaussian Splatting optimization process by employing Newton-Euler equations, focusing specifically on spatiotemporal variations. This approach makes our rendering more robust to rigid and linkage motion changes, and the results demonstrate our method's ability to handle complex trajectories and motions.

\begin{figure*}[!t]
\centering
\includegraphics[width=0.96\textwidth,height=0.19\textheight]{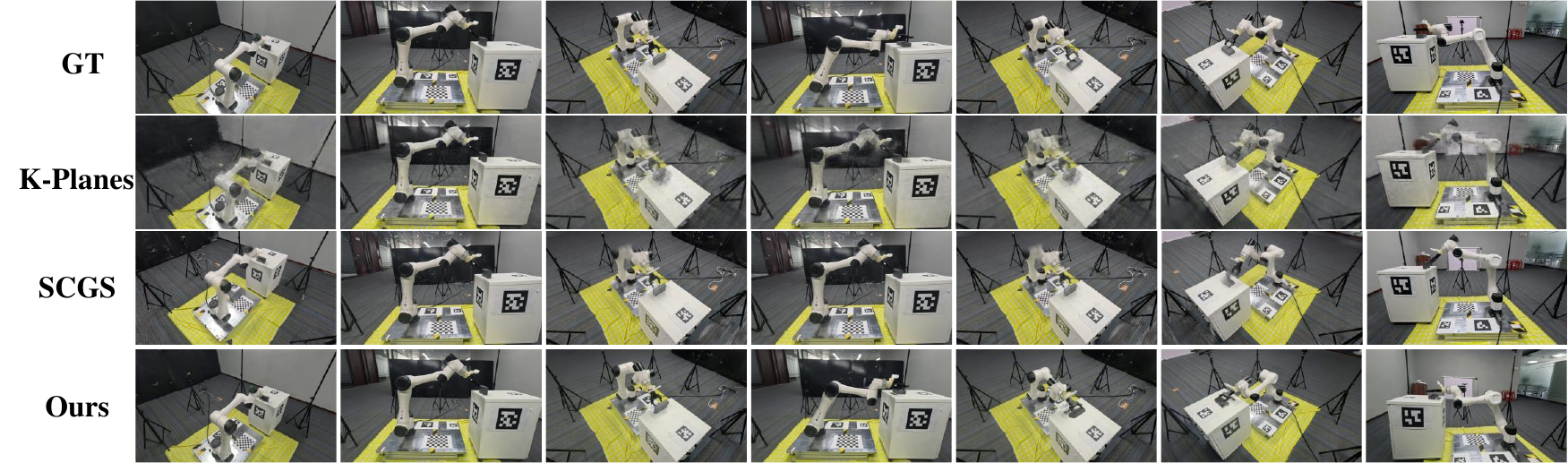}%
\caption{Qualitative result for Push-box}
\label{pushbox}
\end{figure*}

For a dynamic robotic manipulation scene simulator, rendering the manipulated object is equally crucial. Fig. \ref{pushbox} illustrates the reconstruction of a robotic arm pushing a box. As shown, both KPlanes and SC-GS fail to accurately capture the dynamic motion of the robotic arm and the rigid body, consistent with the findings from Robo360~\cite{c64}. In contrast, our method successfully preserves motion and geometric consistency during robotic manipulation tasks.

\subsection{Mesh Quality:}
\begin{figure}[!t]
\centering
\includegraphics[width=0.45\textwidth]{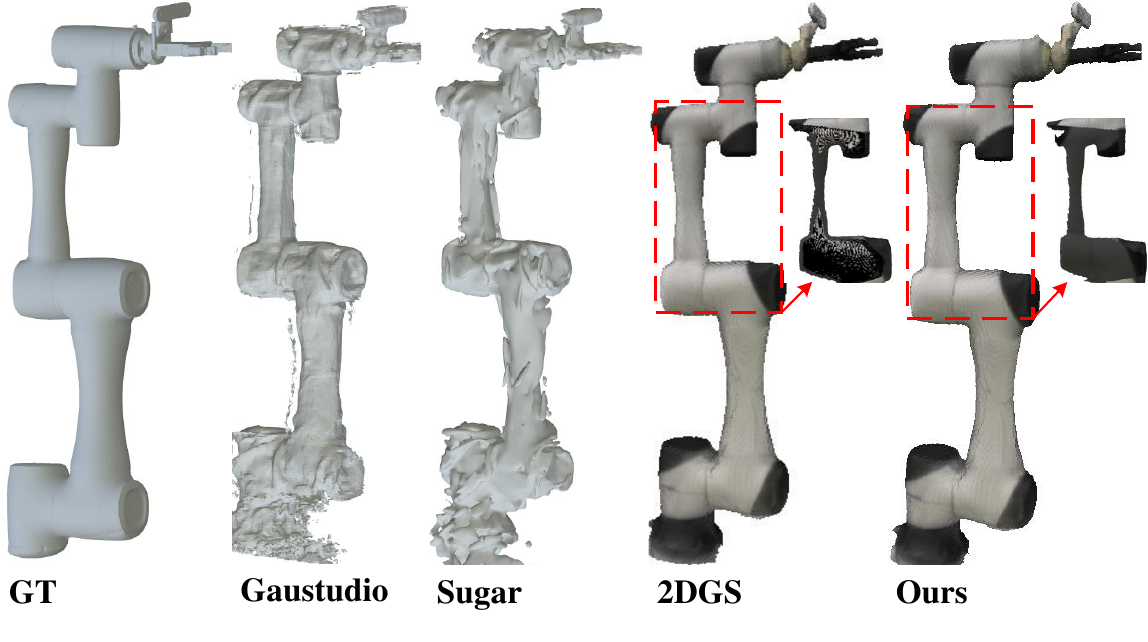}%
\caption{Qualitative result for mesh reconstruction, and the white inner shell of 2DGS mesh is raised by the TSDF sampling strategy, whereas our mesh solves this issue.}
\label{mesh reconstrucion}
\vspace{-3mm}
\end{figure}

We compare our method against SOTA video-based mesh reconstruction approaches, including 2DGS~\cite{c30}, Gaustudio~\cite{c15}, SUGAR~\cite{c27}, as well as ground truth data obtained from a commercial 3D scanner. As shown in Fig.\ref{mesh reconstrucion}, compared to SUGAR and Gaustudio, our approach results in better mesh quality with texture.
Mean Square Error (MSE) and F-score, with a tolerance of 1e-5, were employed to evaluate reconstruction precision and spatial consistency, respectively. Building on 2DGS for our mesh reconstruction, the surface quality remains largely consistent. However, the suboptimal opacity optimization in 2DGS can lead to the inner shell within the robotic arm's mesh, adversely affecting the quantitative results (Table \ref{tab:mesh_ablation}). This issue also compromises collision detection, reducing the realism of the simulation. Our mesh-cleaning technique eliminates redundant meshes generated by TSDF, enhancing alignment and sampling and yielding improved quantitative outcomes. Therefore, our model effectively reconstructs the collision groups of various small modules of the robotic arm in the URDF implementation.


\begin{table}[!t]
\caption{Mesh reconstruction ablation study}
\centering
\label{tab:mesh_ablation}
\begin{tabular}{@{}ccc@{}}
\toprule
\makebox[0.12\textwidth][c]{Methods} & \makebox[0.07\textwidth][c]{MSE} & \makebox[0.12\textwidth][c]{F-score} \\ \midrule
Gaustudio & 0.153 & 0.268 \\
            Sugar& 0.159 & 0.375 \\
            2DGS& 0.153 & 0.228 \\
            Ours& \textbf{0.151} & \textbf{0.721} \\
\bottomrule
\end{tabular}
\end{table}


\subsection{Manipulable Robotic Model:}
In our approach, the simulation generates the movement of objects and the robotic arm, resulting in simulated trajectories for each object and robotic linkages. These trajectories are then transformed into the pre-timestamp global transformation matrix. We enable editing in the Isaac simulation backend~\cite{c6} and seamless rendering in Dynamic Gaussian Splatting (details are shown in the attached video). Additionally, our approach also supports novel-policy editing~\cite{c73}. Novel-policy allows training a position-based policy, distilling a set of trajectory of end-effector pose from policy~\cite{c63} based on the position of object from the reconstruction of our digital asset with rendering and simulation in our method. We currently support custom trajectory editing in Isaac Gym and policy implementation, demonstrating the ability for Sim2Real and Sim2Render applications based on this paradigm (details are shown on part \emph{Policy Implementation} of our webpage). 

\section{Conclusion}

Our goal was to develop a robust Real2Sim framework that significantly reduces the gap between real-world robotic manipulation tasks and their simulated counterparts. We accomplished this by introducing a hybrid representation model that integrates mesh geometry, Gaussian kernels, and physics attributes. This approach ensures high-quality, realistic, and physics-consistent rendering of robotic arm manipulation scenes. Our model was trained and validated across CR~\cite{c8} and UR~\cite{c9} product, demonstrating its effectiveness in constructing accurate URDFs from video data. This method not only enhances the fidelity of simulated environments but also generalizes well to other robotic applications, advancing the state-of-the-art in robotic learning and control. Our current grasping approach uses a position-based policy. However, Gaussian kernels serves as a world representation, enabling accurate rendering from any in-scene camera pose~\cite{c71}. Our engine also supports vision-based grasping policies using our assets and model, allowing for the setup of the render camera and providing precise rendering.

\bibliographystyle{ieeetr} 
\bibliography{ref}

\end{document}